\documentclass[letterpaper, 10 pt, conference]{ieeeconf}  

\usepackage[T1]{fontenc}

\IEEEoverridecommandlockouts                              

\usepackage{amssymb}  
\usepackage{amsmath} 
\usepackage{url}
\usepackage{wrapfig}
\usepackage{graphicx}
\usepackage{subcaption}

\usepackage[dvipsnames]{xcolor}
\usepackage{tikz}

\usepackage{siunitx}
\usepackage{xspace}
\usepackage{tabularx}
\usepackage{booktabs}
\usepackage{balance}
\usepackage{arydshln}  
\usepackage[bookmarks=true,colorlinks=true,citecolor=blue,linkcolor=blue,urlcolor=blue]{hyperref}
\usepackage{float}
\definecolor{steelblue}{HTML}{4682B4}
\newif\ifshowrevisions
\showrevisionstrue

\usepackage{dblfloatfix}
\usepackage{multirow,graphicx,array}

\newif\ifshowcomments
\showcommentstrue
\showcommentsfalse 
\ifshowcomments
    \usepackage[textsize=scriptsize]{todonotes}
    \newcommand{\fix}[1]{{\color{red} #1}}
\else
    \usepackage[disable,textsize=scriptsize]{todonotes}
    \newcommand{\fix}[1]{}
\fi

\makeatletter
\let\NAT@parse\undefined
\makeatother
\usepackage[square, numbers, sort]{natbib}

\title{\LARGE \bf
Atomic-WAM: Execution-Grounded Atomic-Motion \\
Reasoning for World Action Models
}

\author{%
Shijun Shi$^{1\ast \dagger}$,
Zhe Liu$^{2\ast}$,
Zhi Hou$^{3}$,
Yinan Chen$^{1\dagger}$, 
Tao Zhang$^{1}$, \\
Kai Hu$^{4\ddagger}$, 
Yong Liu$^{1}$,
Jiangning Zhang$^{1\ddagger}$
\thanks{$^{1}$Zhejiang University. $^{\dagger}$Work completed during an internship at ACE Robotics.}%
\thanks{$^{2}$University of Hong Kong. $^{\ast}$Equal contribution.}%
\thanks{$^{3}$ACE Robotics.}%
\thanks{$^{4}$Jiangnan University.}%
\thanks{$^{\ddagger}$ Equal advising.}%
}

\begin{document}
\bstctlcite{atomicIEEEcontrol}

\maketitle
\thispagestyle{empty}
\pagestyle{empty}

\begin{abstract}

World Action Models (WAMs) typically adapt video foundation models to robotic control through joint video and action prediction.
However, task-level instructions may leave the next physical interaction underspecified, requiring semantic grounding and state-dependent motion selection.
To this end, we introduce Atomic-WAM, an understand-then-execute framework that connects task reasoning to action generation through atomic-motion descriptions.
Specifically, a vision-language reasoner describes the next local motion from the task instruction and visual context, guiding a WAM to predict future video and an action chunk.
Furthermore, an automatic annotation pipeline curates temporally aligned, contact-enriched supervision from multi-view DROID demonstrations.
To make atomic-motion predictions executable by the WAM, we post-train the reasoner with Execution-Grounded GRPO (EG-GRPO). With the WAM fixed, EG-GRPO adapts the reasoner through rewards for motion appropriateness and execution success.
Finally, Atomic-WAM achieves 39.6\% task success under default instructions and 31.6\% under vague instructions in a real-to-sim evaluation on RoboLab. \\
\noindent
\textbf{Project page:}
\href{https://ssj9596.github.io/Atomic-WAM-project/}{https://ssj9596.github.io/Atomic-WAM-project}

\end{abstract}

\section{Introduction} \label{sec:introduction}

\begin{figure}[!t]
    \centering
    \includegraphics[width=\columnwidth]{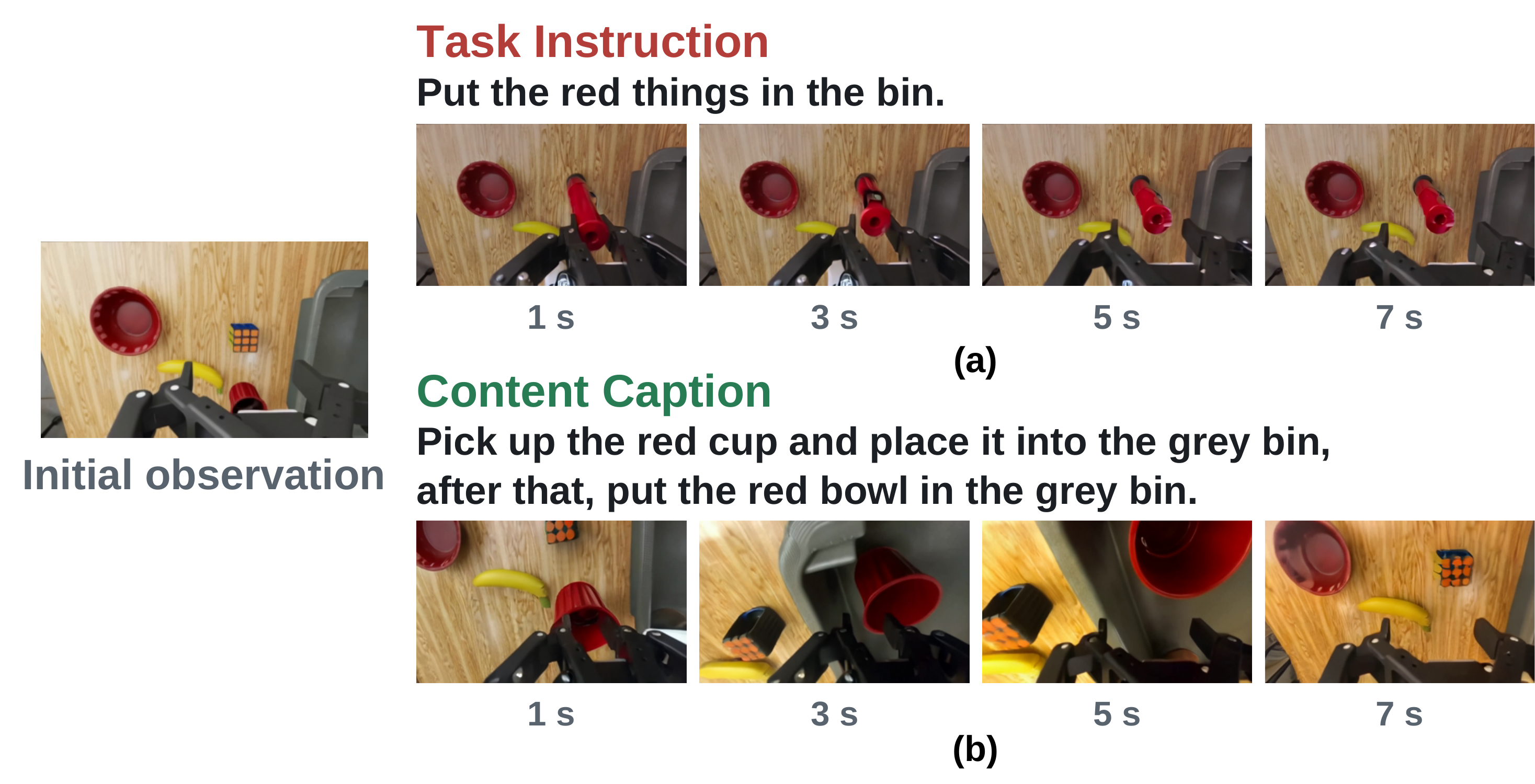}
    \vspace{-0.5cm}
    \caption{\textbf{Language specificity shapes predicted futures.} Given the same initial observation and video backbone, we compare video predictions conditioned on (a) an abstract task instruction and (b) a concrete multi-step description.
    By explicitly specifying the target objects and action sequence, the latter produces a more coherent predicted manipulation sequence in this example.}
    \label{fig:conditioning}
    \vspace{-0.5cm}
\end{figure}

World Action Models (WAMs) transfer visual and motion priors from video foundation models to robotic control by jointly predicting future visual states and actions~\cite{pai2025mimic,ye2026world,li2026causal,kim2026cosmos,agarwal2026cosmos3}.
Through video pretraining, these models acquire rich dynamic priors for modeling physically plausible future evolution. 
However, robot control further requires determining which local evolution is appropriate under the current task and state. Given a fixed task-level instruction, the object to manipulate, the current interaction phase, and the local motion to execute are often not explicitly specified.
Such task-level conditioning leaves two related requirements to the executor.
First, a semantic grounding gap arises when categories, relations, or desired outcomes must be translated into concrete object interactions.
For example, \emph{Put the red things in the bin} requires identifying the relevant objects and deciding how to manipulate them.
Second, a temporal granularity gap arises because a task instruction can remain unchanged as execution progresses through approach, alignment, grasp, transport, and release~\cite{wallwm}.
Descriptive video pretraining does not by itself establish how to resolve these decisions in closed-loop control.
Figure~\ref{fig:conditioning} illustrates the sensitivity of video predictions to language specificity, motivating a more explicit interface between task reasoning and motion execution.

To address these requirements, we introduce \textbf{Atomic-WAM}, an \emph{understand-then-execute} framework that connects task reasoning to video--action generation through an atomic-motion language interface. Specifically, a vision-language model (VLM) reasoner uses the task instruction and visual context to describe the next local interaction, including its target and relevant spatial grounding.
A WAM then predicts the corresponding future video and action chunk. The reasoner revises its description as execution changes the scene, while the task-level objective remains fixed.

Learning such an interface requires temporally aligned atomic-motion supervision, which is rarely available in existing robot demonstrations.
We therefore build an automatic pipeline that partitions the DROID dataset~\cite{khazatsky2024droid} into semantically coherent atomic motions. The interface further allows us to improve grasp learning. To this end, we enrich the corresponding descriptions with visible contact details.
The resulting labels supervise both motion selection and execution at the same atomic-motion granularity.

Nevertheless, supervised atomic-motion learning alone does not guarantee that the selected motion is compatible with the learned executor~\cite{finevla,chen2026steerable}. Existing works introduce intermediate commands, subgoals, or semantic events to decouple high-level reasoning from low-level control~\cite{chen2026steerable,intelligence2026pi,wallwm}, yet such intermediate decision policies are typically not optimized directly from the downstream executor. We therefore introduce execution-grounded reasoner adaptation through EG-GRPO, which closes the loop between the reasoner and the executor: the reasoner selects atomic motions, the executor produces physical outcomes, and these outcomes serve as the learning signal to improve atomic-motion selection.


Building on these advances, Atomic-WAM achieves competitive performance in an out-of-distribution real-to-sim evaluation on RoboLab~\cite{yang2026robolab}, with 39.6\% task success under default instructions and 31.6\% under vague instructions.

In summary, our contributions are threefold:
\begin{itemize}
    \item We introduce \textbf{Atomic-WAM}, an understand-then-execute framework that connects high-level task reasoning to joint video--action generation through state-dependent atomic-motion descriptions.

    \item We develop a \textbf{scalable atomic-motion annotation pipeline} that integrates displacement and gripper cues, multi-view evidence, and visual verification to provide temporally aligned, contact-enriched supervision for both motion selection and execution.

    \item We introduce \textbf{execution-grounded reasoner adaptation} via EG-GRPO, using physical outcomes from a fixed executor to improve atomic-motion selection through closed-loop optimization.
\end{itemize}

\section{Related Work}
\label{sec:related_work}

\subsection{Embodied foundation models.}
Recent embodied policies have benefited from two forms of large-scale pretraining: vision--language pretraining for vision-language-action (VLA) models and video generative pretraining for world action models (WAMs).
VLA models transfer pretrained vision--language representations to action prediction~\cite{brohan2023rt,kim2024openvla,black2024pi_0,shukor2025smolvla,bjorck2025gr00t,team2025gemini,wen2025dexvla}.
Such models benefit from rich semantic priors for objects, scenes, and language-specified tasks, supporting generalization across diverse manipulation tasks.
However, vision--language pretraining primarily emphasizes visual--semantic alignment rather than action-conditioned temporal dynamics.
In contrast, WAMs leverage video generation to model environment dynamics and couple future visual prediction with robot action generation~\cite{ye2026world,agarwal2026cosmos3,li2026causal,kim2026cosmos,bharadhwaj2024gen2act,won2025dual,zhou2025act2goal,liao2026genie,hu2024video,wang2026unified}.
Despite this strength, the textual supervision of video foundation models is largely descriptive and offers limited grounding for semantically specified robot tasks.
Atomic-WAM integrates the strengths of both forms of pretraining: a VLM reasoner progressively translates high-level task semantics into atomic motions, while the WAM predicts the physical evolution and robot actions.

\begin{figure}[t]
 \centering
 \includegraphics[width=\columnwidth]{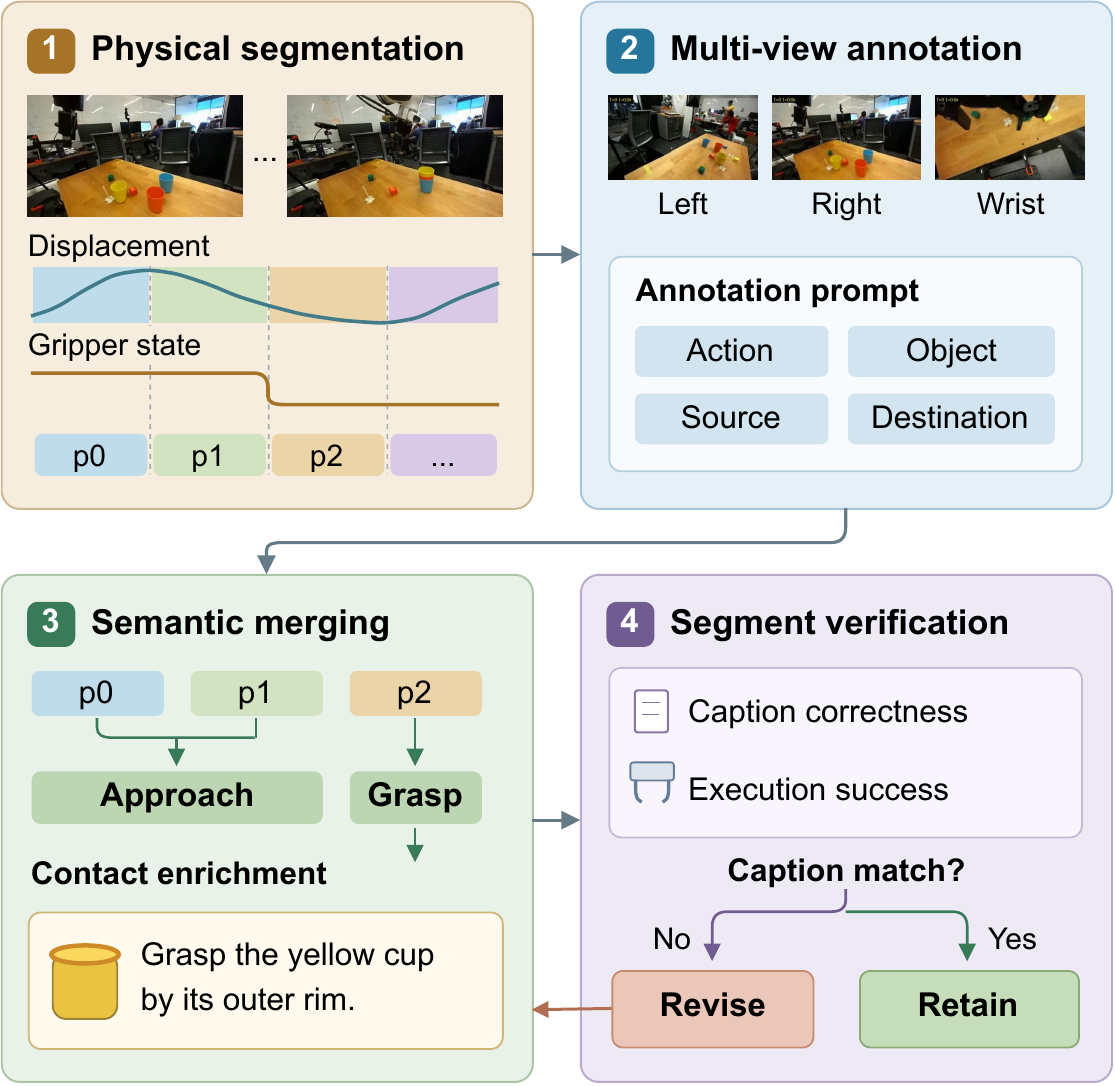}
 \caption{\textbf{Atomic-motion data curation.} Displacement and gripper events first divide a demonstration into physical clips. We then annotate each clip from three views, then use temporal context to merge adjacent clips and refine their descriptions. Visible grasp details further enrich grasp-related captions. Finally, we check caption correctness and execution success; mismatched captions are revised and checked again.}
 \label{fig:exam_annotation}
 \vspace{-0.5cm}
\end{figure}

\subsection{Embodied reasoning for robot control.}
High-level embodied reasoning enables robots to decompose tasks, assess progress, reason about object affordances, and derive spatially grounded plans from multimodal observations.
Recent VLMs have substantially advanced these capabilities~\cite{sermanet2024robovqa,ji2025robobrain,gong2026ace,team2026ace}, but translating high-level reasoning into controllable robot behavior remains challenging.
A growing line of work therefore grounds reasoning into robot policies through richer intermediate representations.
Embodied Chain-of-Thought (ECoT) augments action prediction with intermediate reasoning over plans, subtasks, motions, and visually grounded features~\cite{zawalski2024robotic}.
Steerable Policies train VLAs to follow commands at multiple levels of abstraction, including subtasks, atomic motions, and spatially grounded instructions~\cite{chen2026steerable}.
Similarly, $\pi_{0.7}$ conditions robot behavior on diverse language, metadata, and visual subgoals~\cite{intelligence2026pi}, while WALL-WM organizes video--action learning around semantically coherent action events~\cite{wallwm}.
HarnessWAM further incorporates execution evidence into task management, using progress feedback to support replanning and failure recovery~\cite{gu2026harnesswam}.
These approaches strengthen the interface between reasoning and execution, but execution outcomes are generally not used to directly optimize the high-level reasoner under a downstream policy.
Atomic-WAM closes this learning loop by executing candidate atomic motions with a frozen WAM and updating the reasoner from group-relative rewards that capture both motion appropriateness and execution success.

\begin{figure*}[t]
 \centering
 \includegraphics[width=\textwidth]{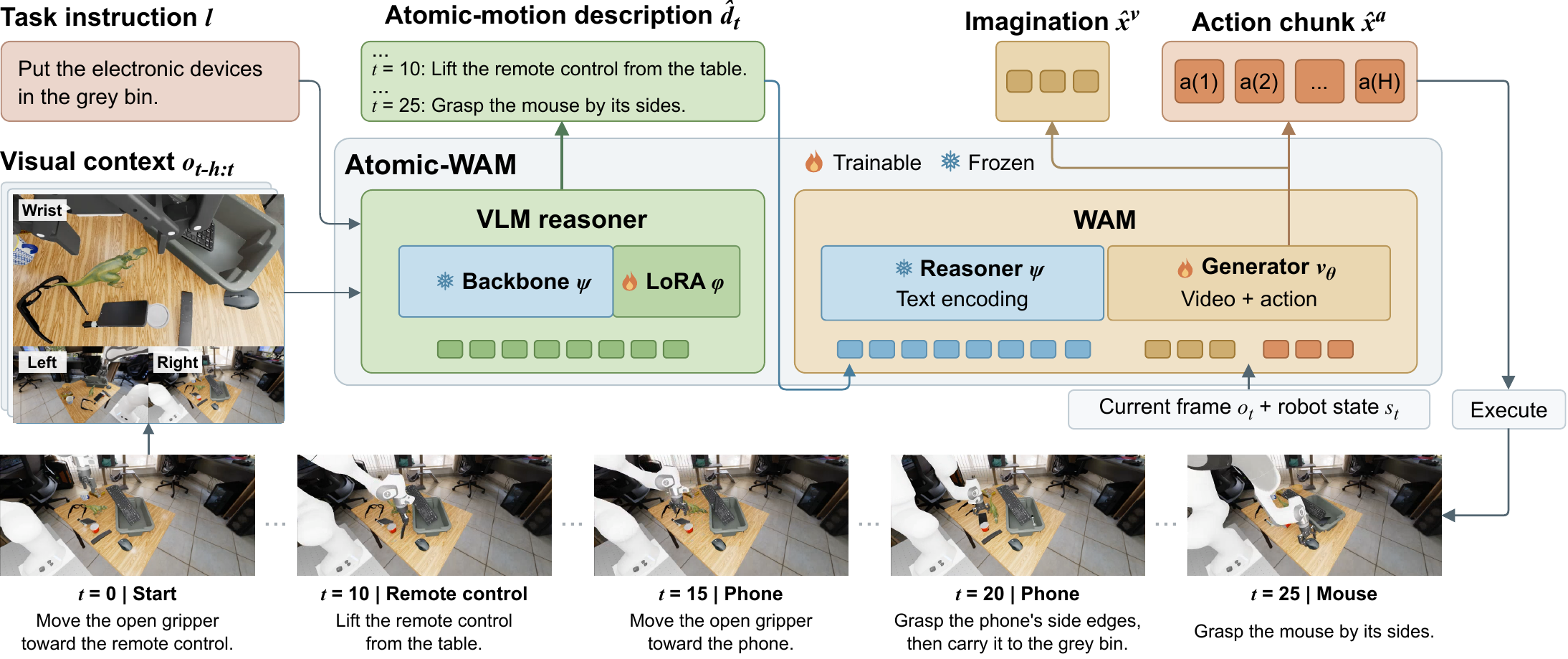}
 \vspace{-0.5cm}
 \caption{\textbf{Atomic-WAM overview}. At decision step $t$, the reasoner predicts an atomic-motion description $\hat{d}_t$ from the task instruction $l$ and visual context $o_{t-h:t}$, where each multi-view observation is formed by concatenating the wrist, left, and right camera views. The WAM conditions on $\hat{d}_t$, $o_t$, and $s_t$ to generate video latents $\hat{x}^v$ and actions $\hat{x}^a$. The lower frames show selected decision steps from a full-task rollout, beginning at \textit{Start} ($t=0$).}
 \label{fig:exam_method}
 \vspace{-0.6cm}
\end{figure*}

\subsection{Robot simulation evaluation.}
Real-world robot evaluation is difficult to reproduce because physical resets and environmental conditions vary across trials~\cite{li2024evaluating}.
By contrast, simulation offers repeatable task instances and controlled variations in scenes and instructions.
While conventional simulation benchmarks support scalable evaluation across diverse manipulation tasks~\cite{liu2023libero,fei2025libero,chen2025robotwin,nasiriany2024robocasa}, substantial training--evaluation overlap can lead to benchmark saturation and provide limited evidence of out-of-distribution generalization.
Recent real-to-sim benchmarks instead evaluate policies trained on real-world data directly in simulation, providing a controlled setting for policy evaluation~\cite{li2024evaluating,yang2026robolab}.
Within this setting, SIMPLER studies whether simulated evaluation reflects real-world policy behavior, while RoboLab further targets out-of-distribution task generalization with 120 novel tasks spanning visual, relational, and procedural competencies
In this work, we adopt RoboLab to evaluate Atomic-WAM under this out-of-distribution real-to-sim setting.

\section{Method}
\label{sec:approach}

Atomic-WAM connects task reasoning to video-action generation through state-dependent atomic-motion descriptions.
We first construct the supervised data from robot demonstrations (Sec.~\ref{sec:atomic_annotation}).
Next, we present the “understand-then-execute” framework and training process (Sec.~\ref{sec:exam_training}).
Finally, we introduce a closed-loop optimization stage that leverages execution feedback to further improve atomic-motion selection (Sec.~\ref{sec:execution_rl}).

\subsection{Atomic-Motion Data Curation}
\label{sec:atomic_annotation}
Robot demonstrations typically provide only task-level instructions, leaving local motions unspecified.
Collecting such supervision densely is expensive: neighboring observations often correspond to the same manipulation phase, making frame-wise annotation redundant.
To reduce this redundancy, we curate atomic-motion annotations bottom-up from fine-grained physical clips, as shown in Fig.~\ref{fig:exam_annotation}.


\noindent\textbf{Motion-grounded segmentation.} Inspired by the speed-based segmentation of human hand trajectories in VITRA~\cite{li2025scalable}, we first derive candidate boundaries from changes in end-effector velocity. Additionally, we incorporate gripper-state transitions to capture interaction phases that are not evident from velocity changes alone. Together, these signals partition each demonstration into physical clips with locally consistent motion and gripper behavior.

\begin{figure*}[t]
 \centering
 \includegraphics[width=\textwidth]{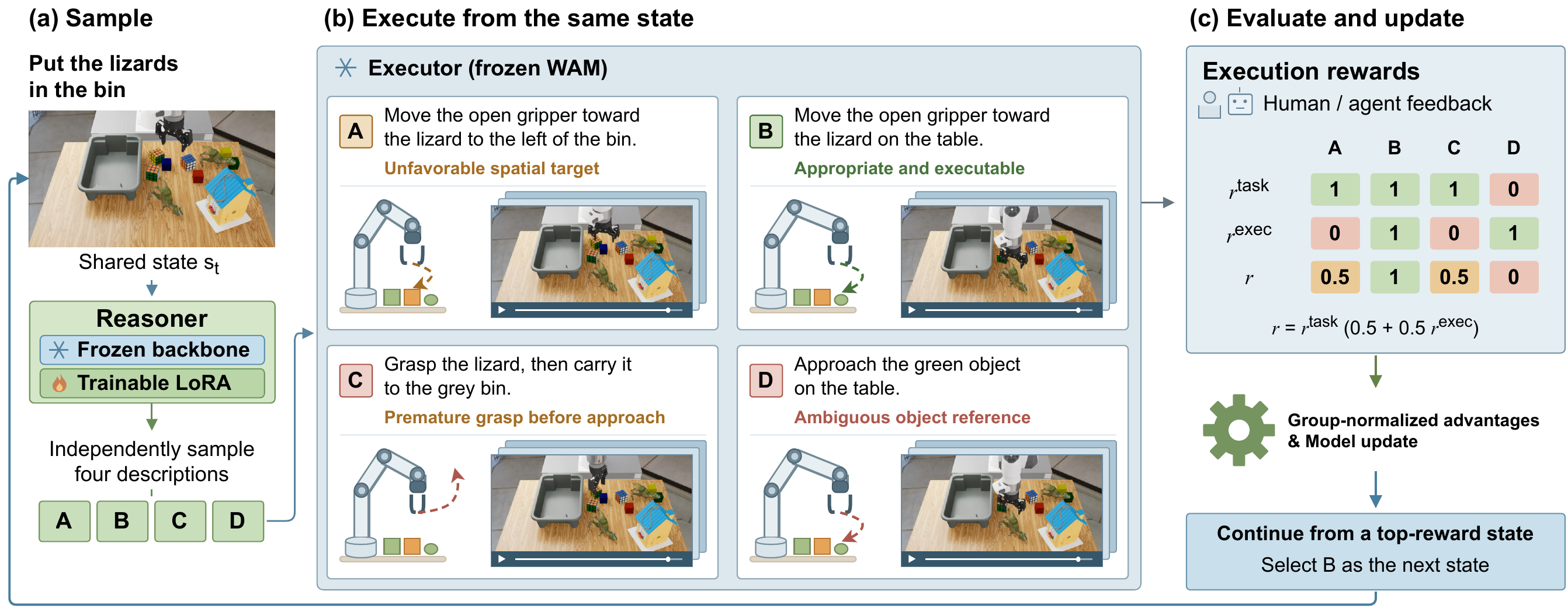}
    \vspace{-0.5cm}
    \caption{\textbf{Closed-loop execution-grounded reasoner training.}
    The reasoner samples four candidate atomic motions from a shared observation, while the frozen WAM executes them independently.
    An evaluator scores motion appropriateness and execution success, and the resulting group-normalized advantages update the reasoner.
    After each sampling round, one of the highest-reward endpoints is randomly selected as the starting state for the next round.}
 \label{fig:reasoner_rl}
 \vspace{-0.6cm}
\end{figure*}

\noindent\textbf{Atomic-motion labeling.}
We build a multi-stage annotation pipeline using models from Qwen family~\cite{yang2025qwen3}.
For each physical clip, we annotate the synchronized left, right, and wrist views separately.
Each annotation records the visible action, manipulated object, source location, and destination.
However, individual physical clips may fragment a complete action, while early clips can yield ambiguous source or destination labels.
We therefore merge adjacent clips into atomic-motion segments using the three-view annotations, boundary observations, and temporal context.
The atomic-motion granularity allows us to capture finer interaction details.
For grasp-related segments, we further enrich the descriptions with visible grasp points and gripper positions.
Finally, we verify each segment's caption for correctness and assess execution success, revising and rechecking mismatched captions. Overall, our pipeline yields 369k atomic-motion segments from 57k episodes, with a 96.8\% checker pass rate using Qwen3.8-Max. We further conduct a manual review of 200 checker-passed samples, obtaining a 90.5\% agreement rate with human judgments.

\subsection{Supervised Understand-then-Execute Learning}
\label{sec:exam_training}
\noindent\textbf{Understand-then-Execute Pipeline.}
Atomic-WAM instantiates the understand-then-execute framework with Cosmos3-Nano~\cite{agarwal2026cosmos3}. As shown in Fig.~\ref{fig:exam_method}, it consists of an autoregressive reasoner and a diffusion-based video--action generator.
The reasoner takes the task instruction $l$ and visual context $o_{t-h:t}$ to predict the next atomic-motion description $\hat d_t$, which then serves as the language condition for WAM together with current observation $o_t$ and robot state $s_t$.
The WAM produces future video latents $\hat x_t^v$ and an action chunk $\hat x_t^a$; once the action chunk is executed, the resulting observation is incorporated into the visual context for the next atomic-motion prediction.

\noindent\textbf{Supervised Learning.}
Atomic-WAM supports two learning objectives within a single architecture.
The atomic-motion annotations from Sec.~\ref{sec:atomic_annotation} supervise both components: each description $d_t$ serves as the prediction target for the reasoner and the language condition for the WAM.
Let $\psi$ and $\theta$ denote the base reasoner and WAM generator parameters, respectively, and let $\phi$ denote the reasoner LoRA parameters. We train the reasoner to predict $d_t$ from the task instruction $l$ and visual context $o_{t-h:t}$:
\begin{equation}
\mathcal{L}_{\mathrm{reason}}(\phi)
=-\mathbb{E}\sum_{j=1}^{|d_t|}
\log p_{\psi,\phi}(d_{t,j}\mid d_{t,<j},o_{t-h:t},l),
\label{eq:reason_loss}
\end{equation}
where $j$ indexes description tokens and $h$ denotes the observation history span. Setting $h=0$ uses only the current observation.
For the WAM, we use joint flow matching~\cite{lipman2022flow} to predict future video latents $x^v$ and action chunks $x^a$ from the current observation \(o_t\), state \(s_t\), and description \(d_t\).
For each modality $m\in\{v,a\}$, we interpolate the target $x^m$ with Gaussian noise:
\begin{equation}
x^m_{\sigma_m}=(1-\sigma_m)x^m+\sigma_m\epsilon^m,
\qquad u^m=\epsilon^m-x^m,
\label{eq:flow_path}
\end{equation}
where $\epsilon^m\sim\mathcal{N}(0,I)$ and $\sigma_m\in[0,1]$.
Let $x_{\boldsymbol{\sigma}}=(x^v_{\sigma_v},x^a_{\sigma_a})$ collect the noisy targets at levels $\boldsymbol{\sigma}=(\sigma_v,\sigma_a)$. For each modality, the WAM predicts a velocity \(v^m_{\psi,\theta}\):
\begin{equation}
\mathcal{L}_{\mathrm{FM}}^m(\theta)
=\mathbb{E}\!\left\|
v^m_{\psi,\theta}(x_{\boldsymbol{\sigma}},\boldsymbol{\sigma};o_t,s_t,d_t)-u^m\right\|^2.
\label{eq:modality_flow_loss}
\end{equation}
Overall objective combines video and action prediction:
\begin{equation}
\mathcal{L}_{\mathrm{exec}}(\theta)
=w_v\mathcal{L}_{\mathrm{FM}}^v(\theta)
+w_a\mathcal{L}_{\mathrm{FM}}^a(\theta),
\label{eq:exec_loss}
\end{equation}
with modality weights \(w_v\) and \(w_a\).

We train the two components in parallel using separate forward passes and optimizers. The reasoner updates LoRA adapter $\phi$, while the WAM updates $\theta$ using $d_t$ encoded by the frozen base reasoner $\psi$ with the LoRA disabled.

\subsection{Closed-Loop Execution-Grounded Reasoner Adaptation}
\label{sec:execution_rl}
\label{sec:test_time_refinement}
Supervised learning provides Atomic-WAM with an initial understand-then-execute capability, but the reasoner is optimized without execution feedback.
Therefore, a motion prediction that appears reasonable from a local perspective may not be poorly realized by the downstream executor.
For example, a prematurely predicted \emph{approach-then-grasp} motion may cause the gripper to close before contact, while a premature \emph{carry-then-release} prediction may release the object too early; likewise, a spatial target can be geometrically valid yet unfavorable for execution.
Moreover, execution errors can move the robot into states where the motion favored by supervised learning is no longer appropriate. We therefore further post-train the reasoner with Execution-Grounded Group-Relative Policy Optimization (EG-GRPO), using physical outcomes from the frozen WAM to optimize atomic-motion reasoning. 
Motivated by recent robot self-improvement methods that learn from interaction-derived experience~\cite{intelligence2025pi,yang2026rise}, we perform this training on a limited set of RoboLab tasks without task-success labels.
The overall training procedure is illustrated in Fig.~\ref{fig:reasoner_rl}.

\begin{table*}[!t]
    \centering
    \caption{
    RoboLab-120 results under vague and default instructions, evaluated over ten runs per task.
    SR denotes task success rate (\%), and Score denotes subtask progress $\times 100$.
    The episode counts shown in parentheses; the best and second-best results are shown in bold and underline, respectively; ties share the same marking.
    Baseline results follow the official RoboLab leaderboard (accessed September 10, 2026). $^{\diamond}$ denotes entries with closed-source weights.
    We denote Atomic-WAM as VLM + WAM to highlight its explicit use of vision--language reasoning, while the model retains a unified architecture.
    }
    \label{tab:robolab-main}
    \setlength{\tabcolsep}{3pt}
    \renewcommand{\arraystretch}{1.1}
    \footnotesize
    \begin{tabular*}{\textwidth}{@{\extracolsep{\fill}}ll*{8}{c}@{}}
        \toprule
        & & \multicolumn{8}{c}{\textbf{Vague instructions}} \\
        \cmidrule(lr){3-10}
        Model & Type & \multicolumn{2}{c}{Relational (420)} & \multicolumn{2}{c}{Visual (840)} & \multicolumn{2}{c}{Procedural (220)} & \multicolumn{2}{c}{Overall (1,200)} \\
        & & SR & Score & SR & Score & SR & Score & SR & Score \\
        \midrule
        Cosmos3-Nano-Policy~\cite{agarwal2026cosmos3} & WAM & 32.4 & 46.4 & 13.5 & 26.1 & \underline{13.6} & 36.5 & 20.6 & 33.2 \\
        Phoenix$^{\diamond}$ & TAMP+FM & -- & -- & -- & -- & -- & -- & -- & -- \\
        BiMind v0.1$^{\diamond}$ & VLA & \underline{36.7} & \textbf{53.0} & \underline{24.5} & \underline{39.8} & \textbf{15.9} & 28.8 & \underline{30.2} & \underline{44.9} \\
        $\pi_{0.5}$~\cite{intelligence2025pi_} & VLA & 24.0 & 38.8 & 12.3 & 25.0 & 10.9 & 30.6 & 15.2 & 28.1 \\
        DreamZero~\cite{ye2026world} & WAM & 19.0 & 32.3 & 9.5 & 22.0 & \textbf{15.9} & \underline{37.1} & 14.9 & 27.0 \\
        Cosmos3-Edge-Policy~\cite{agarwal2026cosmos3} & WAM & 23.3 & -- & 11.5 & -- & 11.8 & -- & 15.4 & -- \\
        $\pi_0$-FAST~\cite{pertsch2025fast} & VLA & 18.8 & 29.4 & 5.2 & 11.6 & 3.2 & 14.2 & 9.2 & 16.8 \\
        GR00T N1.6~\cite{bjorck2025gr00t} & VLA & 4.8 & 19.6 & 4.3 & 12.3 & 4.1 & 15.2 & 5.4 & 14.0 \\
        $\pi_0$~\cite{black2024pi_0} & VLA & 6.2 & 18.5 & 0.8 & 5.2 & 1.4 & 6.8 & 2.8 & 8.5 \\
        \midrule
        \textbf{Atomic-WAM (ours)} & \textbf{VLM + WAM} & \textbf{38.6} & \underline{50.0} & \textbf{26.3} & \textbf{42.1} & \textbf{15.9} & \textbf{38.2} & \textbf{31.6} & \textbf{46.7} \\
        \midrule
        & & \multicolumn{8}{c}{\textbf{Default instructions}} \\
        \cmidrule(lr){3-10}
        Model & Type & \multicolumn{2}{c}{Relational (420)} & \multicolumn{2}{c}{Visual (840)} & \multicolumn{2}{c}{Procedural (220)} & \multicolumn{2}{c}{Overall (1,200)} \\
        & & SR & Score & SR & Score & SR & Score & SR & Score \\
        \midrule
        Cosmos3-Nano-Policy~\cite{agarwal2026cosmos3} & WAM & \textbf{49.3} & \textbf{60.9} & 26.1 & 42.9 & \textbf{27.3} & \textbf{46.8} & \underline{36.8} & \underline{51.9} \\
        Phoenix$^{\diamond}$ & TAMP+FM & 25.0 & 37.3 & \underline{33.6} & \underline{45.4} & 21.8 & 30.3 & 34.4 & 45.9 \\
        BiMind v0.1$^{\diamond}$ & VLA & 35.7 & 51.7 & 29.4 & 43.0 & 21.4 & 33.3 & 33.3 & 47.2 \\
        $\pi_{0.5}$~\cite{intelligence2025pi_} & VLA & 33.8 & 46.8 & 23.8 & 38.8 & 21.8 & 41.6 & 28.0 & 43.4 \\
        DreamZero~\cite{ye2026world} & WAM & 33.3 & 45.5 & 17.5 & 31.9 & \underline{26.4} & 42.3 & 25.7 & 39.8 \\
        Cosmos3-Edge-Policy~\cite{agarwal2026cosmos3} & WAM & 26.7 & -- & 16.8 & -- & 15.0 & -- & 22.9 & -- \\
        $\pi_0$-FAST~\cite{pertsch2025fast} & VLA & 23.3 & 36.7 & 8.8 & 19.5 & 6.4 & 16.4 & 15.5 & 26.9 \\
        GR00T N1.6~\cite{bjorck2025gr00t} & VLA & 10.0 & 23.2 & 5.0 & 15.1 & 2.7 & 11.1 & 7.2 & 17.1 \\
        $\pi_0$~\cite{black2024pi_0} & VLA & 6.4 & 18.8 & 1.9 & 8.2 & 0.5 & 6.6 & 5.0 & 12.2 \\
        \midrule
        \textbf{Atomic-WAM (ours)} & \textbf{VLM + WAM} & \underline{45.0} & \underline{53.9} & \textbf{34.0} & \textbf{49.2} & 25.9 & \underline{45.7} & \textbf{39.6} & \textbf{53.7} \\
        \bottomrule
    \end{tabular*}
    \vspace{-0.5cm}
\end{table*}

\noindent\textbf{RL Formulation.}
We formulate the reasoner as a high-level policy $\pi_{\psi,\phi}$ over atomic-motion descriptions.
Given the task instruction $l$ and visual context $o_{t-h:t}$ at decision step $t$, we sample a group of $K$ candidate motions from the behavior policy, denoted as $\{\hat d_{t,i}\}_{i=1}^{K}$.
The executor realizes each candidate independently from the same initial state, producing local rollout $\xi_{t,i}$.
We assign each candidate an execution-grounded reward
\begin{equation}
r_{t,i}
=
R(\hat d_{t,i},\xi_{t,i}),
\label{eq:eg_reward}
\end{equation}
as detailed below.
The rewards are normalized within the group to obtain the advantage
\begin{equation}
\widehat A_{t,i}
=
\frac{
r_{t,i}
-
\operatorname{mean}\{r_{t,1},\ldots,r_{t,K}\}
}{
\operatorname{std}\{r_{t,1},\ldots,r_{t,K}\}
+\delta
},
\label{eq:grpo_advantage}
\end{equation}
where $\delta$ is a small constant for numerical stability.
This group-wise comparison evaluates alternative motion selections from the same state.
After each group is evaluated, we randomly select one endpoint among the highest-reward candidates as the next decision state, from which the reasoner samples a new candidate group.

\noindent\textbf{Execution-Grounded Reward.}
The reward consists of two parts.
Task reward $r^{\mathrm{task}}_{t,i}$ evaluates whether the predicted motion $\hat d_{t,i}$ is an appropriate next step toward completing task $l$ from the current state. Given the task instruction, current observation, and predicted atomic motion, the evaluator assigns a score of 1 if the motion is appropriate and 0 otherwise.
Execution reward $r^{\mathrm{exec}}_{t,i}$ is used to evaluate whether the downstream executor has successfully executed $\hat d_{t,i}$.
The evaluator examines the resulting rollout $\xi_{t,i}$; if the specified motion is successfully completed, it assigns a value of 1; otherwise, it assigns a value of 0.
We combine the two components hierarchically:
\begin{equation}
r_{t,i}
=
r^{\mathrm{task}}_{t,i}
\left(
0.5 + 0.5\,r^{\mathrm{exec}}_{t,i}
\right).
\label{eq:reward_components}
\end{equation}
Therefore, inappropriate motions will receive a zero reward regardless of whether they are executed, while appropriate motions will receive the full reward only if they are successfully executed.

\noindent\textbf{Optimization Objective.}
We update the reasoner LoRA $\phi$ using the GRPO objective~\cite{shao2024deepseekmath} with the group-normalized advantages $\widehat A_{t,i}$:
{\small
\begin{equation}
\begin{aligned}
\mathcal{J}_{\mathrm{GRPO}}(\phi)
=
\frac{1}{K}\sum_{i=1}^{K}
\Big[
&\min\!\left(
s_{t,i}\widehat A_{t,i},
\operatorname{clip}(s_{t,i},1-\epsilon,1+\epsilon)\widehat A_{t,i}
\right) \\
&-\beta D_{\mathrm{KL}}
\!\left(
\pi_{\psi,\phi}\,\|\,\pi_{\mathrm{ref}}
\right)
\Big].
\end{aligned}
\label{eq:grpo_objective}
\end{equation}
}
where
\begin{equation}
s_{t,i}
=
\frac{
\pi_{\psi,\phi}
(\hat d_{t,i}\mid o_{t-h:t},l)
}{
\pi_{\psi,\phi_{\mathrm{old}}}
(\hat d_{t,i}\mid o_{t-h:t},l)
}.
\end{equation}
Here, $\pi_{\mathrm{ref}}$ denotes the frozen supervised reasoner, while $\epsilon$ and $\beta$ control policy clipping and KL regularization, respectively.
Only the LoRA parameters $\phi$ are updated; the base reasoner $\psi$ and downstream executor remain frozen.

\section{Experiments}
\label{sec:experiments}

\subsection{Experiment Setup}
\label{sec:experimental_setup}

\noindent\textbf{Training Data.}
Following the official Cosmos3-Nano-Policy-DROID setup, we use 57k successful DROID demonstrations and convert them into 369k atomic-motion segments for supervised training.

\noindent\textbf{Evaluation.}
We evaluate each of the 120 RoboLab tasks~\cite{yang2026robolab} over 10 random seeds.
The benchmark covers three overlapping competencies: \emph{visual} grounding of color, semantics, and size; \emph{procedural} competence in affordances, reorientation, and stacking; and \emph{relational} competence in conjunctions, counting, and spatial relationships.
For each task, we evaluate both default and vague instructions, with the latter leaving more details implicit and placing greater demands on semantic grounding.
We report task success rate (SR) and the benchmark's subtask-progress score, following the official evaluation protocol.

\noindent\textbf{Implementation Details.}
We initialize the reasoner and WAM from Cosmos3-Nano~\cite{agarwal2026cosmos3}. Training is conducted on 32 NVIDIA A800 GPUs.
Each visual input combines a $360{\times}640$ wrist view with two $180{\times}320$ side views.
The WAM is conditioned only on the current observation. It is trained for 14{,}000 steps with batch size 768 and a learning rate of $2\times10^{-4}$, using $w_v=w_a=10$ and 32-step action chunks.
Reasoner training mixes three-observation and single-observation contexts at an $8{:}2$ ratio. The three-observation context contains the current observation together with observations from 8 and 16 frames earlier.
The reasoner is trained for 3{,}000 steps with the same batch size and a learning rate of $5\times10^{-5}$. Its LoRA adapters use $r=128$ and $\alpha=256$.
Both stages use fused AdamW with linear learning-rate decay.
For EG-GRPO, we collect rollouts on four RoboLab tasks: discarding an apple, placing Rubik's cubes in a bin, placing electronic devices in a bin, and packing cans.
We sample four candidate descriptions per state at temperature 1.0, performing 30 updates with batch size 8 and learning rate $10^{-5}$.

\subsection{Main Results}
Table~\ref{tab:robolab-main} compares Atomic-WAM with Cosmos3-Nano-Policy and other RoboLab baselines under default and vague instructions.
Atomic-WAM achieves the best overall performance in both settings, reaching 39.6\% SR and a subtask-progress score of 53.7 under default instructions, and 31.6\% SR and 46.7 under vague instructions.
When moving from default to vague instructions, Atomic-WAM's success rate drops by only 8.0 percentage points, whereas Cosmos3-Nano-Policy's success rate drops by 16.2 percentage points.
This reduced sensitivity to instruction specificity suggests that explicit atomic-motion reasoning helps recover object, relation, and motion information that is otherwise left implicit in the instruction.

The competency-level results further support this trend.
Atomic-WAM shows the most consistent gains on visual tasks, improving success from 26.1\% to 34.0\% under default instructions and from 13.5\% to 26.3\% under vague instructions.
For relational tasks, the benefit is more pronounced under vague instructions, where success increases from 32.4\% to 38.6\%.
In contrast, the gains on procedural tasks are smaller: under default instructions, Atomic-WAM reaches 25.9\% success, slightly below the 27.3\% of Cosmos3-Nano-Policy.
These results suggest that Atomic-WAM provides the clearest advantage when successful execution requires resolving information left implicit in the instruction, while tasks that depend more heavily on procedural execution remain constrained by the underlying WAM.

\subsection{Analysis and Ablations}

\begin{figure}[t]
\centering
\includegraphics[width=\columnwidth]{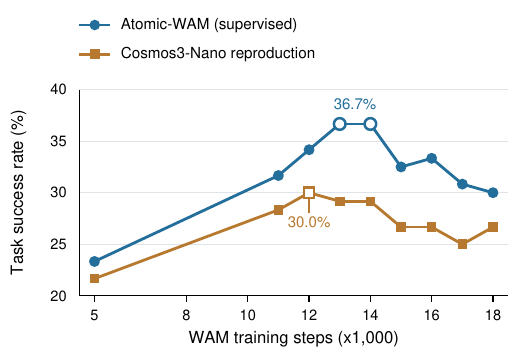}
\vspace{-0.5cm}
\caption{\textbf{RoboLab success rate across training checkpoints.}
    Results are averaged over three runs on RoboLab-120 with default instructions.}
\label{fig:ood_training_duration}
\vspace{-0.2cm}
\end{figure}

\noindent\textbf{Training efficiency.}
The official Cosmos3-Nano-Policy is trained for 10{,}000 steps with a global batch size of 8{,}192~\cite{agarwal2026cosmos3}.
On our 32 NVIDIA A800 GPUs, reproducing this configuration would require more than 700 hours, making iterative experimentation impractical.
We therefore reduce the global batch size to 768 and train for up to 20{,}000 steps, bringing the training time down to approximately 80 hours.
With this configuration, our Cosmos3-Nano-Policy reproduction achieves 30.0\% success, compared with 36.8\% for the published checkpoint.
Figure~\ref{fig:ood_training_duration} tracks performance across training checkpoints.
The reproduced baseline reaches its best result of 30.0\% after 12{,}000 steps, whereas the supervised Atomic-WAM attains 36.7\% after both 13{,}000 and 14{,}000 steps.
We also find that further training does not consistently improve performance: at 18{,}000 steps, the success rates drop to 26.7\% and 30.0\%, respectively.

\noindent\textbf{Ablation on atomic-motion supervision.}
Table~\ref{tab:ablation} examines the contribution of atomic-motion supervision.
Introducing the understand-then-execute framework with atomic-motion descriptions improves task success from 30.0\% to 32.5\%.
Further incorporating visible grasp points and gripper-position details raises success to 36.7\%.
This demonstrates that contact-level information can further enhance the effectiveness of atomic motion supervision, as more precise grasping positions and gripper states reduce spatial ambiguity in grasping actions.

\begin{table}[t]
    \centering
    \caption{\textbf{Ablation of atomic-motion supervision.}
    Results are averaged over three runs on RoboLab-120 with default instructions.}
    \label{tab:ablation}
    \small
    \setlength{\tabcolsep}{6pt}
    \renewcommand{\arraystretch}{1.08}
    \begin{tabular}{ccc}
        \toprule
        Atomic motion & Contact & SR (\%) \\
        \midrule
         &  & 30.0 \\
        \checkmark &  & 32.5 \\
        \checkmark & \checkmark & \textbf{36.7} \\
        \bottomrule
    \end{tabular}
    \vspace{-0.5cm}
\end{table}

\begin{table}[t]
    \centering
    \caption{\textbf{Ablation of reasoner adaptation.}
    Results are averaged over three runs on RoboLab-120 with default instructions.}
    \label{tab:reward_ablation}
    \small
    \setlength{\tabcolsep}{5pt}
    \renewcommand{\arraystretch}{1.08}
    \begin{tabular}{l c}
        \toprule
        Reasoner adaptation & SR (\%) \\
        \midrule
        Supervised reasoner & 36.7 \\
        GRPO w/o execution reward & 34.2 \\
        EG-GRPO & \textbf{39.2} \\
        \bottomrule
    \end{tabular}
    \vspace{-0.5cm}
\end{table}

\noindent\textbf{Analysis of closed-loop reasoner adaptation.}
We first investigate the impact of stronger high-level reasoning while keeping WAM fixed.
We replace the trained reasoner with Codex agent based on GPT-5.6-Sol and consider two settings: continuing from the resulting state after each execution, and allowing the agent to backtrack and revise its motion plan.
On a subset of 64 RoboLab tasks, the trained reasoner completes 23 tasks, while the Codex agent completes 33 without backtracking and 50 with it.
These results indicate that stronger reasoning capabilities alone provides only limited improvements.
If the selected motion is not executable, even semantically reasonable predictions can lead to unfavorable states that are difficult to recover from.
The larger benefits from backtracking further demonstrate the necessity of execution feedback to adapt action selection to the WAM.

We next evaluate the effect of execution feedback on reasoner adaptation.
Starting from the same supervised initialization, GRPO without the execution reward decreases success from 36.7\% to 34.2\%.
This suggests that motion appropriateness alone is insufficient when the selected motion cannot be reliably executed by the WAM.
In contrast, EG-GRPO additionally incorporates execution success and improves the success rate to 39.2\%.

\noindent\textbf{Limitations and Future Work.}
During EG-GRPO training, we initially use a Codex agent powered by GPT-5.6-Sol to obtain reliable reward signals.
However, computing the execution reward requires assessing every local rollout, making the process costly in both inference latency and token usage.
In our current setting, 30 updates correspond to 960 rollouts, which remain manageable for human evaluation.
We therefore use human judgments of execution success.
Future work will focus on developing an efficient automatic execution evaluator for larger-scale EG-GRPO training.

\section{Conclusion}
We introduced Atomic-WAM, an understand-then-execute framework that translates task-level instructions into state-dependent atomic motions before action generation.
To support this framework, we build a scalable annotation pipeline that provides temporally aligned, contact-enriched atomic-motion supervision.
Building on this supervised stage, we further adapt the reasoner with EG-GRPO, using physical outcomes to improve executable atomic-motion reasoning.
On RoboLab benchmark, Atomic-WAM achieves 39.6\% and 31.6\% success under default and vague instructions, respectively, improving over Cosmos3-Nano-Policy by 2.8 and 11.0 percentage points.
The substantially larger gain under vague instructions shows that Atomic-WAM is particularly effective when task instructions are underspecified.
Overall, Atomic-WAM provides a direct path for translating increasingly capable vision--language reasoning into closed-loop robotic execution.










{\small 
\balance
\bibliographystyle{IEEEtranN}
\bibliography{references}
}

\clearpage

\end{document}